\documentclass{article}

    \PassOptionsToPackage{numbers, sort&compress}{natbib}

 \usepackage[main, final]{neurips_2026}

\usepackage[utf8]{inputenc}
\usepackage[T1]{fontenc}
\usepackage{hyperref}
\usepackage{url}
\usepackage{booktabs}
\usepackage{amsfonts}
\usepackage{nicefrac}
\usepackage{microtype}
\usepackage[dvipsnames]{xcolor}

\usepackage{amsmath}
\usepackage{amssymb}
\usepackage{mathtools}
\usepackage{amsthm}

\usepackage{array}
\usepackage{tabularx}
\usepackage{multirow}
\usepackage{mathtools}
\usepackage{subcaption}
\usepackage{bbm}
\usepackage{wrapfig}

\newcolumntype{L}[1]{>{\raggedright\arraybackslash}p{#1}}
\newcommand{\arrowtag}[1]{\hfill(\(#1\))}

\usepackage[capitalize,noabbrev]{cleveref}

\usepackage[frozencache]{minted}
\definecolor{amber}{rgb}{0.72, 0.53, 0.04}
\usepackage{afterpage}

\usepackage{tikz}
\usetikzlibrary{shapes.geometric}

\title{Simple Baselines are Competitive with Code Evolution}

\definecolor{mydarkblue}{rgb}{0,0.08,0.45}

\hypersetup{
    pdftitle={},
    pdfauthor={},
    pdfsubject={},
    pdfkeywords={},
    pdfborder=0 0 0,
    pdfpagemode=UseNone,
    colorlinks=true,
    linkcolor=mydarkblue,
    citecolor=mydarkblue,
    filecolor=mydarkblue,
    urlcolor=mydarkblue,
    pdfview=FitH
}

\author{%
  Yonatan Gideoni\thanks{Work partially done during an internship at Sakana AI. Code is available at \href{https://github.com/YonatanGideoni/code_evo_simple_baselines}{\url{https://github.com/YonatanGideoni/code_evo_simple_baselines}}.} \\
  University of Oxford\\
  \texttt{yg@robots.ox.ac.uk} \\
  \And
  Sebastian Risi \\
  Sakana AI and ITU Copenhagen \\
  \And
  Yarin Gal \\
  University of Oxford
}

\begin{document}

\maketitle

\begin{abstract}
    Code evolution is a family of techniques that rely on large language models to search through possible computer programs by evolving existing code. While code evolution pipelines have shown impressive performance across domains, many are highly complex and are typically not compared to simpler alternatives. To prevent bad comparisons from stagnating the field, we propose two simple baselines and compare them to popular pipelines across several domains: finding better mathematical bounds, designing agentic scaffolds, developing heuristic combinatorial optimization algorithms, and machine learning competitions. Surprisingly, the baselines match or outperform existing methods across all domains, prompting a closer look at which factors drive performance in different settings. For mathematical bounds, we find that the search space matters far more than the exact search algorithm, with both simple and complex methods performing similarly given the same search space. Thus, expanding the search space is what matters for improving performance, and does so across all pipelines. Moreover, different domain knowledge embedded in prompts can significantly affect the search's efficiency, potentially leading to overestimating a pipeline's sample efficiency. For automated agentic scaffold design, we show that typical high-variance evaluations lead to all automated search methods picking subpar scaffolds relative to a simple majority vote. To mitigate this, we propose improved evaluation procedures that reduce stochasticity while keeping the search economically feasible. Overall, these results indicate that further improving code evolution's performance depends more on the setup surrounding the search pipeline than on the pipeline itself. We hope these insights will enable developing new code evolution methods and help spur meaningful progress in the field.
\end{abstract}

\section{Introduction}
For many problems, a solution can take the form of a computer program, thus allowing automated program search to solve a variety of tasks. This has been demonstrated across several domains: for mathematical and algorithmic discovery, \citet{novikov2025alphaevolve} find programs that improve on several mathematical bounds. \citet{hu2024automated} use LLMs to design programs defining agentic scaffolds which solve math and science problems. \citet{chan2024mle} test how well LLM-based automated coding methods can perform in machine learning competitions.

Many systems search over code space by feeding programs into language models and asking them to evolve, crossover, and recombine them, thereby building LLM pipelines that do code evolution. However, these pipelines consist of numerous design choices, including using ensembles of language models for diversity and cost efficiency, automated parent program selection to maximize diversity, and various ways of receiving feedback \citep{novikov2025alphaevolve, openevolve, lange2025shinkaevolve}. These many design choices are often not ablated, nor are the pipelines compared to simpler alternatives.

Historically, bad comparisons have threatened to stagnate various subfields. In interpretability, \citet{adebayo2018sanity} showed that many existing image saliency methods give similar saliency maps to edge detectors and are independent of the model or the data. Studying drivers of LLM reasoning capabilities, \citet{shao2025spurious} find that some perceived improvements were due to using models that benefit even from random rewards. \citet{henderson2018deep} showed various shortcomings in how deep reinforcement learning (RL) was evaluated, such as high stochasticity and different implementations leading the same method to give meaningfully different results. \citet{agarwal2021deep} further demonstrate that reliance on summary statistics across RL runs results in non-robust evaluations, rendering seemingly strong results unreliable.

To prevent code evolution from suffering similar shortcomings, we ask: \textit{what matters for code evolution?} To systematically test this, we propose two simple baselines -- random sampling from the language model and a more sequential version thereof -- and compare them to several existing pipelines over four different domains.\footnote{Although the baselines are also a form of code evolution, ``code evolution'' here generally refers to more complicated methods unless specified otherwise.} Each domain tests how methods perform under a different budget constraint, such as a limited API budget when finding mathematical bounds \citep{novikov2025alphaevolve}, the number of function evaluations when designing agentic scaffolds \citep{hu2024automated} or finding heuristics for combinatorial optimization problems \citep{imajuku2026ale}, or wall-clock time in machine learning competitions \citep{chan2024mle}. All domains are tested using either the standard or above-standard budgets in the literature.

\begin{figure*}[t]
    \centering
    \begin{subfigure}[b]{0.42\linewidth}
        \centering
        \includegraphics[width=\linewidth]{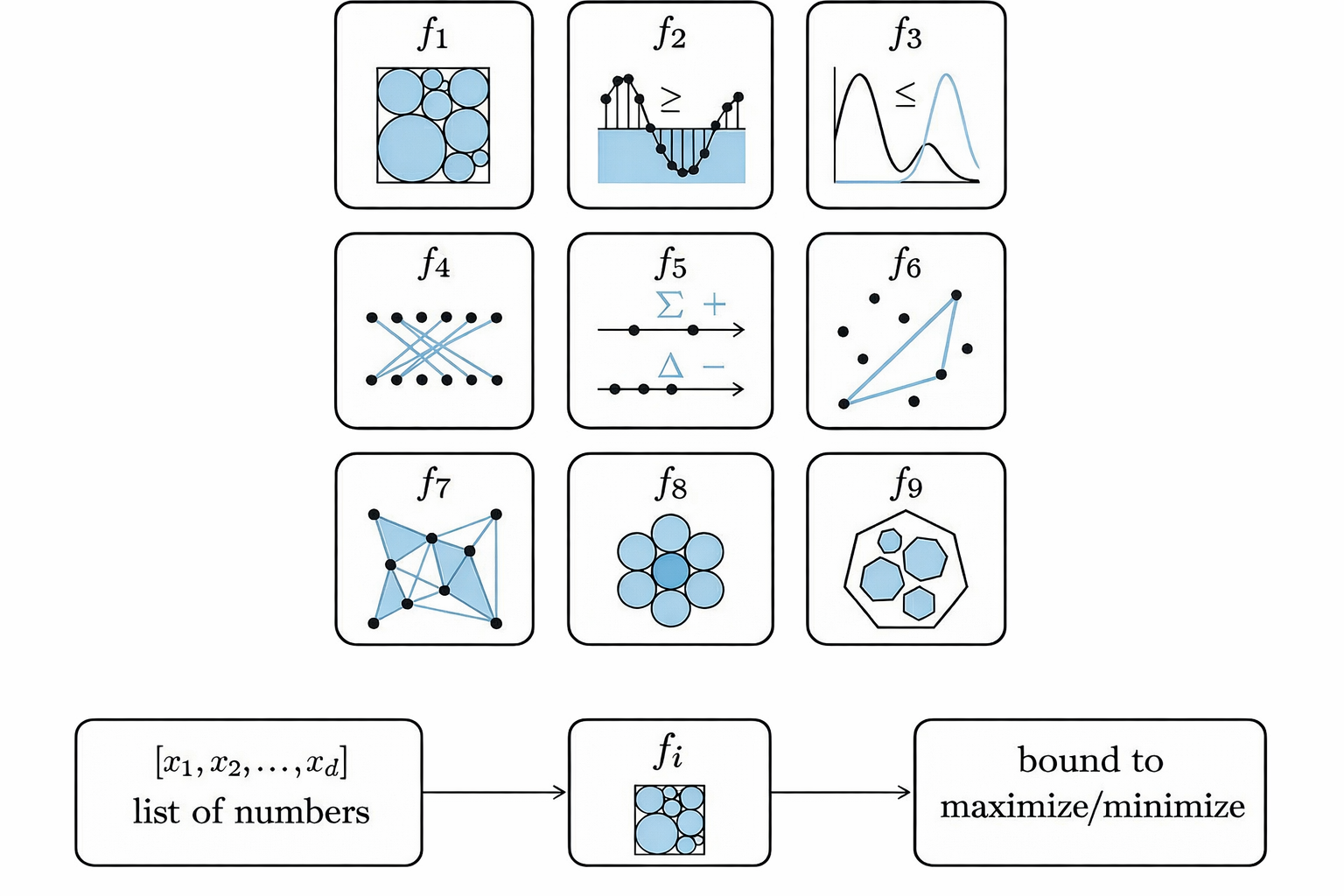}
        \caption{Each problem defines a function that should be maximized/minimized.}
        \label{fig:probs_as_funcs}
        \end{subfigure}\hfill
        \begin{subfigure}[b]{0.22\linewidth}
        \centering
        \includegraphics[
            width=\linewidth,
            trim=150pt 200pt 810pt 0pt,
            clip
        ]{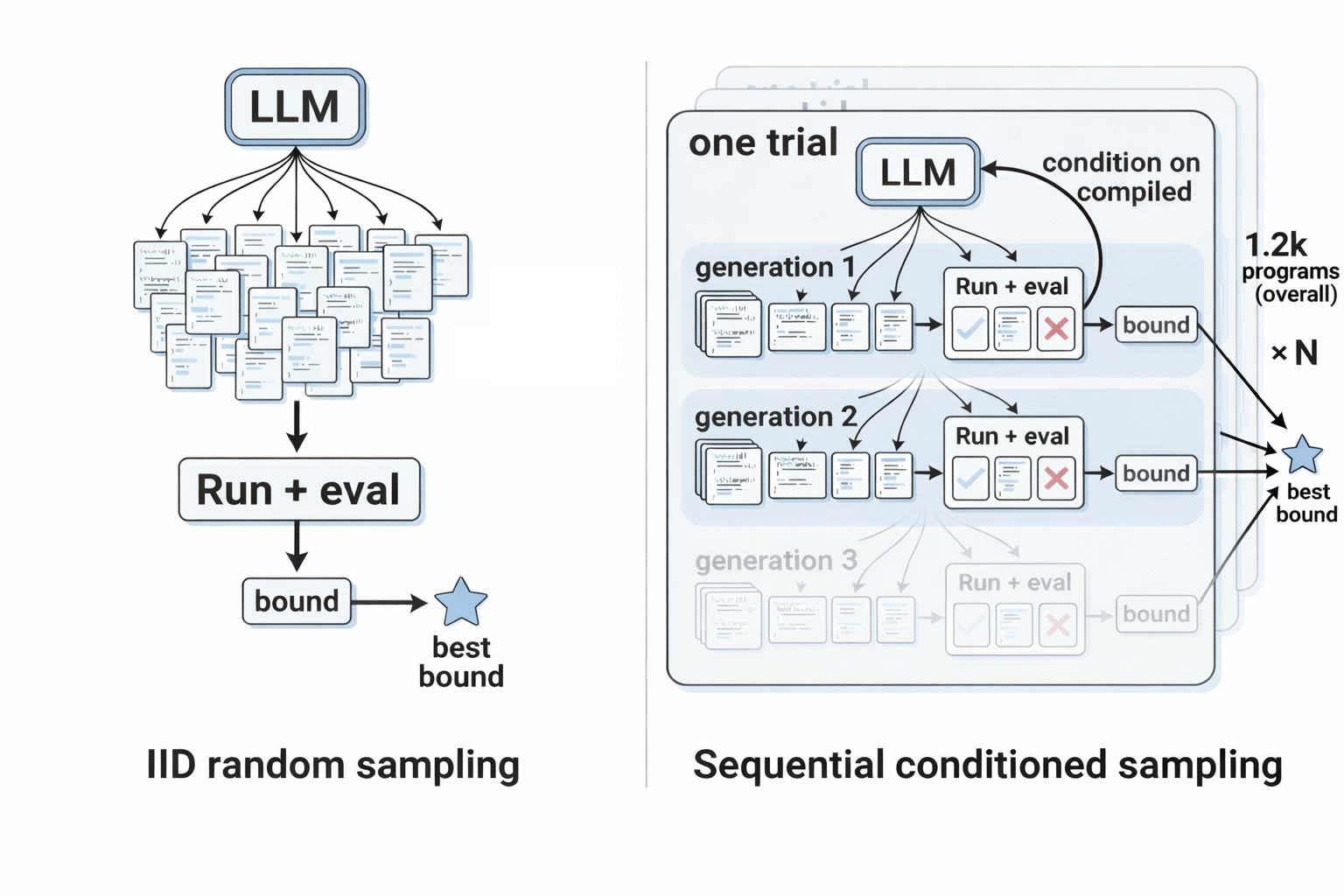}
        \caption{IID random sampling}
        \label{fig:iid_rs}
    \end{subfigure}\hfill
    \begin{subfigure}[b]{0.33\linewidth}
        \centering
        \includegraphics[
            width=\linewidth,
            trim=0pt 0pt 340pt 10pt,
            clip
        ]{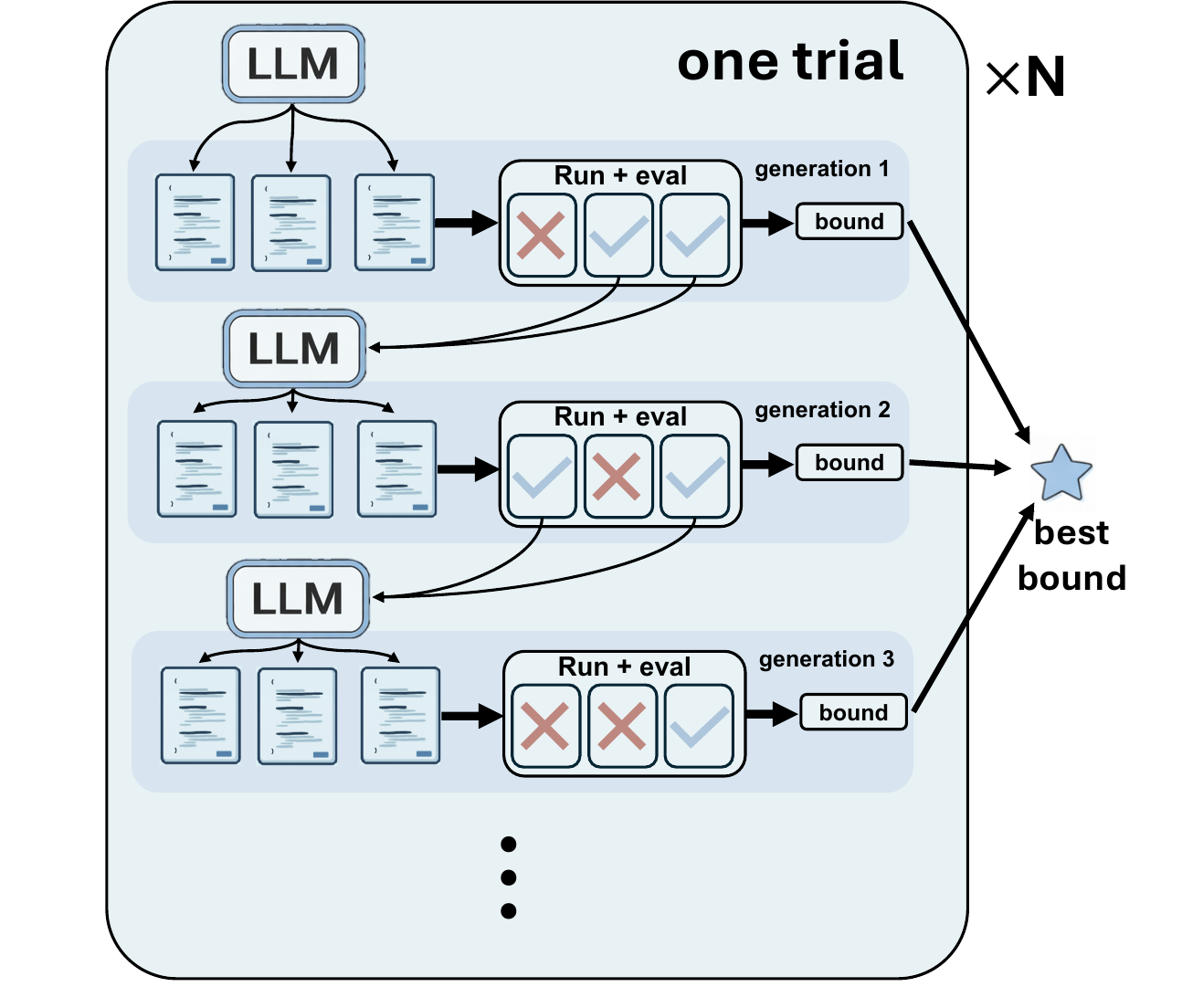}
        \caption{Sequential conditioned sampling} 
        \label{fig:scs}
    \end{subfigure}
    \caption{\textbf{(a)} For AlphaEvolve-style mathematical bounds each problem effectively defines a function that gets as input a list of numbers, possibly with some extra structure, and outputs a mathematical bound that should be maximized or minimized. \textbf{(b, c)} The two baselines, \textbf{(b)} randomly sampling a set of programs IID from an LLM and picking the best one and \textbf{(c)} generating a set of programs, evaluating them, and generating a new set conditioned on some of the programs that ran successfully.
    }
    \label{fig:setup}
    \vspace{-14pt}
\end{figure*}

Surprisingly, in all settings and under the same budget constraints, at least one of the baselines matches or exceeds a purpose-built code evolution pipeline. To understand why, we look into what matters for these search processes. When searching for mathematical bounds, simple and complicated pipelines seem to find high-performing bounds similarly well (§\ref{subsec:mathbounds_res}). Thus, changing the search space -- and thereby the performance ceiling -- improves the discovered bounds, which simple and complex pipelines similarly manage to find. This implies existing code evolution systems perform a confined open-ended search, whereas true open-endedness matters much more for improving performance (§\ref{subsec:search_space}). Moreover, domain knowledge embedded in prompts can substantially alter a search's efficiency, resulting in a pipeline mistakenly seeming more efficient than another (§\ref{subsec:dk}). When searching for agentic scaffolds, we find that all automated search methods suffer from non-robust high-variance summary statistics, leading subpar scaffolds to be selected (§\ref{sec:agent_evo}). To mitigate this, we propose new robust evaluations that reduce stochasticity while keeping the search economical (§\ref{subsec:stoch_red}).

Notably, the search space, domain knowledge, and evaluation methodology are typically not evolved but manually designed by method developers or domain experts. Thus, further performance improvements would likely stem from components external to the search pipeline, which in existing systems are typically kept fixed. We conclude by discussing implications and a set of best practices for future work (§\ref{sec:discuss}).

\vspace{-9pt}
\section{Background}
\vspace{-7pt}
We briefly present code evolution here, with additional related work delegated to Appendix \ref{app:relwork}. Code evolution pipelines typically work by starting with some program and using an LLM to generate new, improved versions of it. Previous programs are often given in the LLM's prompt, selected to maximize a mix of diversity and fitness. This is broadly how systems like AlphaEvolve \citep{novikov2025alphaevolve}, ShinkaEvolve \citep{lange2025shinkaevolve}, OpenEvolve \citep{openevolve}, and ADAS \citep{hu2024automated} work. Some of these systems have many components, often requiring thousands of lines of code.

One example application of code evolution is in finding bounds for the circle packing problem. There, one must maximize the sum of $n$ circle radii while ensuring all circles fit in a unit box with no mutual overlap. Given a list of numbers defining centers and radii, it is straightforward to automatically check whether they define a valid packing, thereby giving a new bound, with this process illustrated in Figure \ref{fig:probs_as_funcs}. Finding a better bound is thus reduced to finding a list of numbers with some properties. Code evolution designs functions that output such lists, either by searching over the numbers and returning the best found candidate or by constructing these lists directly.

Previous works using code evolution for discovery mention how changing a problem's search space or using domain expertise can improve a search's performance and efficiency but do not deeply discuss it. \citet{novikov2025alphaevolve} discuss how occasionally injecting human priors into their code helped them find better matrix multiplication algorithms. \citet{georgiev2025mathematical} present a large-scale case study using AlphaEvolve in conjunction with other tools to find better bounds over 67 problems, noting that a good formulation or telling a pipeline about a theorem can make the difference between finding an improved bound or the pipeline getting stuck.

In some code evolution applications, simple pipelines similar to the baselines presented here are widely used. For ARC AGI, many of the best-performing methods are based on selecting the best program from a large set of generated candidates, perhaps with some additional domain-specific tweaks, e.g. \citet{greenblatt2024getting}. Because we focus on whether the additional complexity in some code evolution systems is beneficial, we do not compare to these simple pipelines.

\vspace{-9pt}
\section{Baselines for Code Evolution}
\vspace{-7pt}
The two baselines are illustrated in Figures \ref{fig:iid_rs} and \ref{fig:scs}. The first baseline, IID random sampling \mbox{(\textbf{IID RS})}, consists of prompting a language model to produce code that solves some task. After evaluating all programs the best candidate is picked, based on the best-found bound for mathematical discovery or on a validation performance for the other problems.

The second baseline is a modification of IID RS, designed to better deal with problems requiring iterative improvements. For example, in the kissing problem, one of the problems tackled in \citet{novikov2025alphaevolve}, one must find the largest possible set of vectors satisfying some property. LLMs often construct the solution directly by specifying these vectors and, at least in their initial answer, do not extend known constructions. Thus iterative methods would likely perform better on similar problems.

Sequential conditioned sampling (\textbf{SCS}) works by first generating a set of programs, akin to IID RS, and then generating another by conditioning on a random subset of the programs that successfully ran in the previous generation. Thus, both baselines have no explicit fitness-based selection. This is repeated for a few generations, and optionally then restarts from scratch. SCS' sequentiality comes at the cost of having each program be more expensive to generate as it results in much longer prompts and is less parallelizable than IID RS. The loss of parallelism is potentially significant in cases where the main bottleneck is wall-clock time and not API cost, function evaluations, or compute.

Operationally, there are a few ways to measure complexity. Typical code evolution systems like ShinkaEvolve have tens of unique hyperparameters, while IID RS has none and SCS has three. This is disregarding shared hyperparameters, like the language model's sampling temperature. Another rough metric for complexity is code length -- both baselines take at most a single file of Python to fully write out, while complex code evolution codebases are often much larger. When we refer to the baselines being ``simple'', operationally we mean simple with respect to these metrics.

\looseness=-1 When comparing methods, different pipelines can arrive at similar solutions or reach the same apparent performance ceiling. We consider two results to be equal if they are the same as per \texttt{numpy.isclose}, with default relative and absolute tolerances, as solutions may differ due to numerical precision or otherwise small numerical instabilities. This is important as for some tasks, like machine learning competitions or when finding mathematical bounds, seemingly small improvements can still be meaningful.

\vspace{-6pt}
\section{Finding Mathematical Bounds}
\vspace{-4pt}
\label{sec:mathbounds}
To test how the baselines compare to code evolution, we test how well they can find mathematical bounds for nine problems from \citet{novikov2025alphaevolve}. These problems are subdivided into those belonging to analysis, combinatorics, or geometry. We sample three, two, and four problems from each category respectively, with the slight imbalance due to combinatorics having only two problems. Brief summaries of the problems and their bounds are in Table \ref{tab:probs} in Appendix \ref{app:probs}, with longer explanations in \citet{novikov2025alphaevolve}.

As AlphaEvolve is closed-source, we compare to a similar open-source sample-efficient pipeline, ShinkaEvolve \citep{lange2025shinkaevolve}. For a fair comparison, all methods use Gemini-2.5 models, so differences in performance are not due to different model families. ShinkaEvolve is restricted to Gemini-2.5 Pro, Flash, and Flash Lite, akin to AlphaEvolve, while both baselines use Gemini-2.5 Pro. Note that the Gemini-2.5 family's knowledge cutoff is January 2025, several months before AlphaEvolve came out, so good performance is not due to memorizing known answers \citep{comanici2025gemini}. Additional technical details are in Appendix \ref{app:math_techdet}.

For all methods reported results are for a \$20 budget. Under these conditions, ShinkaEvolve runs for 500-800 generations, or 25-40 generations per dollar. For reference, the circle packing run in \citet{lange2025shinkaevolve} costs about \$12 and ran for 150 generations (12.5 generations per dollar). Specifically for estimating uncertainties beyond-budget baseline runs were used, see Appendix \ref{app:bootstrap} for details. Only a single ShinkaEvolve run is used per problem as the experiments are very expensive -- running ShinkaEvolve and the two baselines altogether costs \textgreater\$70 per problem. Some of ShinkaEvolve's hyperparameters were slightly tuned to improve its performance, see Appendix \ref{app:math_techdet} for details.

\subsection{Different Methods Should Use the Same Prompts: 
\label{subsec:dk}
Domain Knowledge Significantly Affects Sample Efficiency}
\begin{listing}[h]
\begin{minted}[breaklines,
autogobble,
fontsize=\footnotesize,
frame=single]{markdown}
You are an expert mathematician specializing in circle packing problems and computational geometry. The best known result for the sum of radii when packing 26 circles in a unit square is 2.635.

Key directions to explore:
1. The optimal arrangement likely involves variable-sized circles
2. A pure hexagonal arrangement may not be optimal due to edge effects
3. The densest known circle packings often use a hybrid approach
4. The optimization routine is critically important - simple physics-based models with carefully tuned parameters
5. Consider strategic placement of circles at square corners and edges
6. Adjusting the pattern to place larger circles at the center and smaller at the edges
7. The math literature suggests special arrangements for specific values of n
8. You can use the scipy optimize package (e.g. LP or SLSQP) to optimize the radii given center locations and constraints

Be creative and try to find a new solution better than the best known result.
\end{minted}
\caption{The default circle packing prompt used by ShinkaEvolve. In addition to broadly stating the problem, the prompt contains much domain knowledge which can aid the search. However, competing methods may not have the same domain knowledge.}
\vspace{-6pt}
\label{lst:shinka_cp_prompt}
\end{listing}
Many existing works use different prompts when comparing to other pipelines \citep{lange2025shinkaevolve,wang2025thetaevolve, yuksekgonul2026learning}. However, we find that even the same pipeline can give drastically different results given two different prompts. To illustrate, we run ShinkaEvolve three times on the circle packing problem, using its default prompt with expert hints (see Listing \ref{lst:shinka_cp_prompt}) and with a prompt containing minimal domain knowledge (Appendix \ref{app:min_cp_prompt}). Otherwise, all hyperparameters are kept fixed, including Shinka's default initial program. When using the default expert prompt, Shinka finds the same bound as \citet{lange2025shinkaevolve} 3/3 times, whereas given the minimal domain-knowledge prompt this bound is found in only 1/3 runs. For additional ablations, see Appendix \ref{app:dk_ablate}.

Thus, to ensure fair comparisons, throughout the paper all methods are given the same minimal domain-knowledge prompts unless noted otherwise. Where relevant, pipelines are also given an uninformative initial program.

\subsection{Results}
\begin{table*}[ht]
    \centering
    \caption{Best bounds found using code evolution and baselines. Best results are bolded, second best are underlined. AlphaEvolve results are excluded from comparisons but included for reference as it uses an unknown but likely much higher budget and more domain knowledge. Sequential conditioned sampling (SCS) matches or exceeds ShinkaEvolve on most problems, with IID random sampling (IID RS) also exhibiting competitive performance, occasionally outperforming other methods. Arrows denote whether higher or lower is better. \# problems $\geq$ $M$ is the number of problems for which a method matches or exceeds method $M$. More significant digits are used when discovered bounds are close. Each result is achieved using a \$20 budget.}
    \vspace{-8pt}
    \begin{tabular}{L{3.9cm} >{\color{gray!105}}c c c c}
        & & & \multicolumn{2}{c}{Baselines (ours)} \\
        Problem & {\color{black}AlphaEvolve} & ShinkaEvolve & IID RS & SCS \\
        \midrule
        \# problems $\geq$ ShinkaEvolve & 7/9 &  & 4/9 & 6/9 \\
        \# problems $\geq$ AlphaEvolve &  & 3/9 & 2/9 & 4/9 \\
        Average rank \arrowtag{\downarrow}  &  & 1.9 & 2.3 & 1.8 \\
        \midrule
        First autocorr. ineq. \arrowtag{\downarrow} 
            & 1.505 
            & \underline{1.522}
            & 1.535 
            & \textbf{1.519} \\

        Second autocorr. ineq. \arrowtag{\uparrow} 
            & 0.8962 
            & \textbf{0.8955}
            & 0.8739 
            & 0.8795 \\

        Uncertainty ineq. \arrowtag{\downarrow} 
            & 0.3521 
            & \textbf{0.3521}
            & \textbf{0.3521} 
            & \textbf{0.3521} \\

        Erd\H{o}s' min. overlap \arrowtag{\downarrow} 
            & 0.3809 
            & \textbf{0.3810}
            & \underline{0.3811} 
            & 0.3812 \\

        Sums/differences of sets \arrowtag{\uparrow} 
            & 1.1584 
            & 1.1095
            & \textbf{1.1237} 
            & \underline{1.1178} \\

        Max--min dist. ratio \arrowtag{\downarrow} 
            & 12.88926 
            & \textbf{12.88923}
            & \textbf{12.88923} 
            & \textbf{12.88923} \\

        Heilbronn triangles \arrowtag{\uparrow} 
            & 0.0365 
            & \underline{0.0356}
            & 0.0334 
            & \textbf{0.0365} \\

        Kissing number in 11D \arrowtag{\uparrow} 
            & 593 
            & \underline{402}
            & \textbf{438} 
            & \textbf{438} \\

        Circle packing \arrowtag{\uparrow} 
            & 2.63586 
            & \textbf{2.63598}
            & 2.632 
            & \underline{2.63590} 
    \end{tabular}
    \label{tab:results_math}
    \vspace{-11pt}
\end{table*}

\label{subsec:mathbounds_res}
Table \ref{tab:results_math} shows that both baselines perform well, matching or exceeding Shinka on $4/9$ and $6/9$ problems for IID RS and SCS respectively.\footnote{Using all the baseline-generated programs, beyond the \$20 limit, results in SCS matching ShinkaEvolve on circle packing and finding an improved bound of 1.1216 for sums/differences of sets. IID RS finds a slightly better bound of 1.529 for the first autocorrelation inequality.} Interestingly, SCS matches or exceeds AlphaEvolve on $4/9$ problems as well, in spite of likely using a lower budget and less domain knowledge. Thus, the baselines seem similarly if not more performant than sophisticated code evolution methods.

\begin{wrapfigure}{r}{0.5\textwidth}
    \centering
    \includegraphics[width=\linewidth]{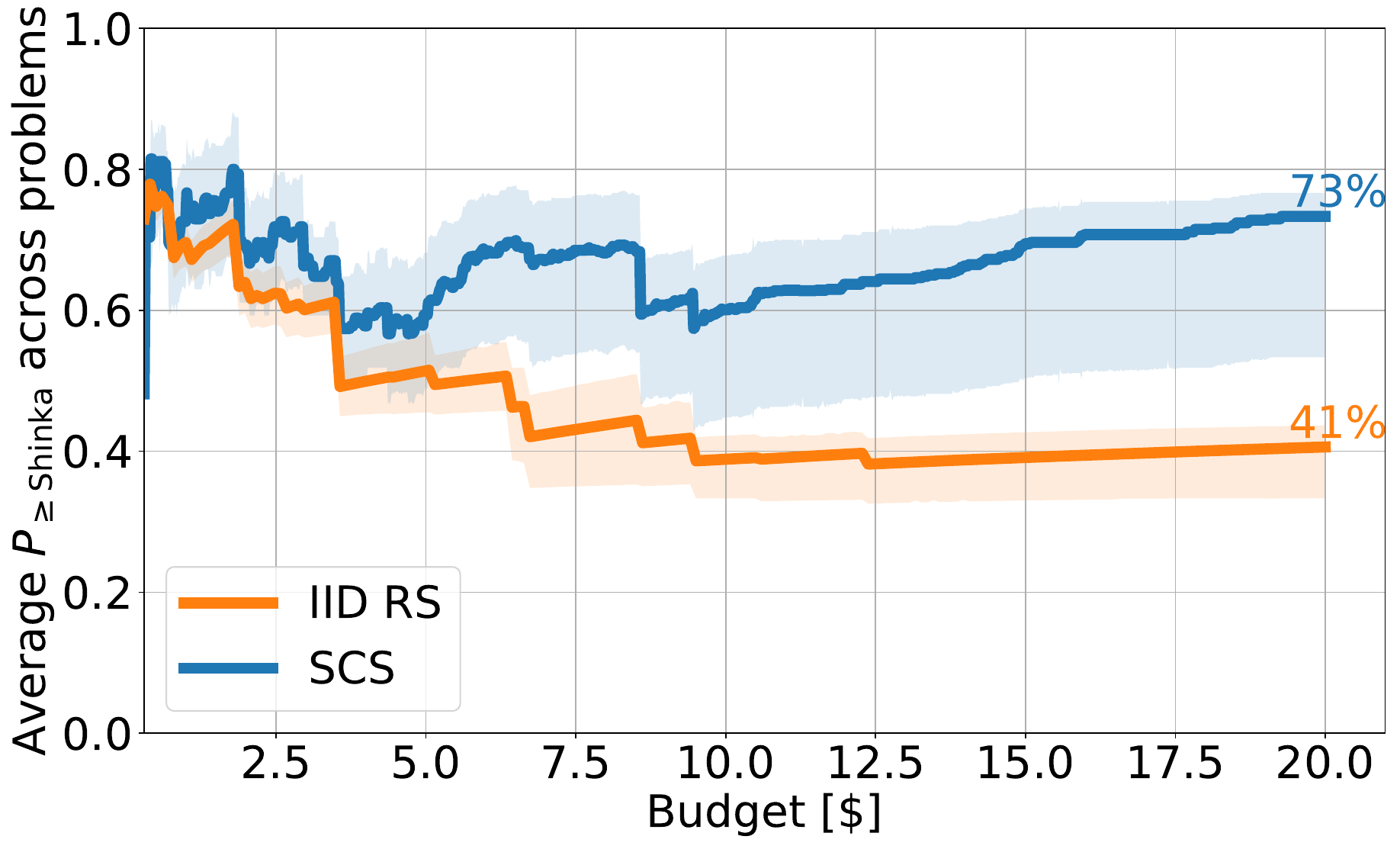}
    \caption{Average probability of matching or exceeding ShinkaEvolve at a given budget over the 9 problems for the two baselines. Colored numbers on the right are the probabilities at the max budget of \$20. Both baselines perform well over different budgets, with sequential conditioned sampling (SCS) generally outperforming ShinkaEvolve. Shaded regions are asymmetric 95\% confidence intervals, see Appendix \ref{app:bootstrap} for details and Appendix \ref{app:per_prob_p_imp} for per-problem breakdowns.}
    \label{fig:prob_imp}
    \vspace{-10pt}
\end{wrapfigure}

To compare the different baselines' cost effectiveness, we calculate the probability that a method matches or exceeds ShinkaEvolve under a given budget. This is done using a bootstrap estimate, with details in Appendix \ref{app:bootstrap}. Figure \ref{fig:prob_imp} demonstrates that both baselines compare favorably to ShinkaEvolve across all budgets. The probabilities initially decrease due to ShinkaEvolve having a warmup period, as it is based on file edits and not full-file generations. Relying on code diffs and other mechanisms, such as occasionally using cheaper LLMs, does not seem to make ShinkaEvolve significantly more cost-efficient.

Here a method's efficiency is measured relative to its API budget, whereas other constraints can be limiting as well. With respect to wall-clock time, the two baselines are much faster due to them being easily parallelizable across CPU cores, requiring 1-3 hours per problem relative to ShinkaEvolve's $\sim$10 hours. In Appendix \ref{app:math_sample_eval_effic} we find that the baselines are sample-efficient also with respect to the number of function evaluations.

\vspace{-7pt}
\subsection{A Problem's Search Space Dictates its Performance Ceiling}
\vspace{-5pt}
\label{subsec:search_space}
As shown in Table \ref{tab:results_math}, all methods find a similar bound for some problems, while still finding similarly performing solutions for others. Thus, given a sufficient budget, all methods can likely reach the same performance ceiling. If this is the case then the only way to improve performance would be by expanding a problem's search space.

To test this, we take one of the problems and improve its formulation. Changing a problem's formulation is reflected by changing the code evolution's verifier and thus the problem's search space. Specifically, we take the uncertainty inequality and improve its formulation so it is easier to optimize and searches over a larger class of functions, with details in Appendix \ref{app:uncert_ineq_form}. Note that AlphaEvolve's formulation is from \citet{gonccalves2017hermite}, with AlphaEvolve and all tested methods here improving the bound from 0.3523 to 0.3521. The new formulation results in all methods -- both baselines and ShinkaEvolve -- finding an improved bound of 0.3482 given the same \$20 budget. Changing the formulation yields a larger improvement than search alone over a given setup, regardless of the search pipeline. We similarly analyze three additional problems in Appendix \ref{app:prob_form_extra_probs}.

Notably, a problem's formulation is designed by domain experts, not the code evolution systems. If good formulations are what make a problem amenable to automated search, with different search methods performing similarly, then many design choices in existing pipelines may be redundant.

The importance of a problem's search space is interesting in regards to \textit{open-endedness}. \citet{hu2024automated, xiong2026learning} and \citet{zhang2026hyperagents} discuss code evolution being theoretically Turing-complete and a form of open-ended search, but how open-ended are current code evolution systems if in practice they do not alter the most important parts of a problem? In this sense, current systems exhibit a confined form of open-endedness, whereas what matters for achieving better performance is altering parts of the setup which are currently kept fixed. Making code evolution more open-ended is an exciting avenue for future work.

\vspace{-9pt}
\section{Evolving Agentic Scaffolds is Highly Stochastic}
\vspace{-5pt}
\label{sec:agent_evo}

\begin{figure*}[!b]
    \vspace{-15pt}
    \centering
    \includegraphics[width=0.99\linewidth]{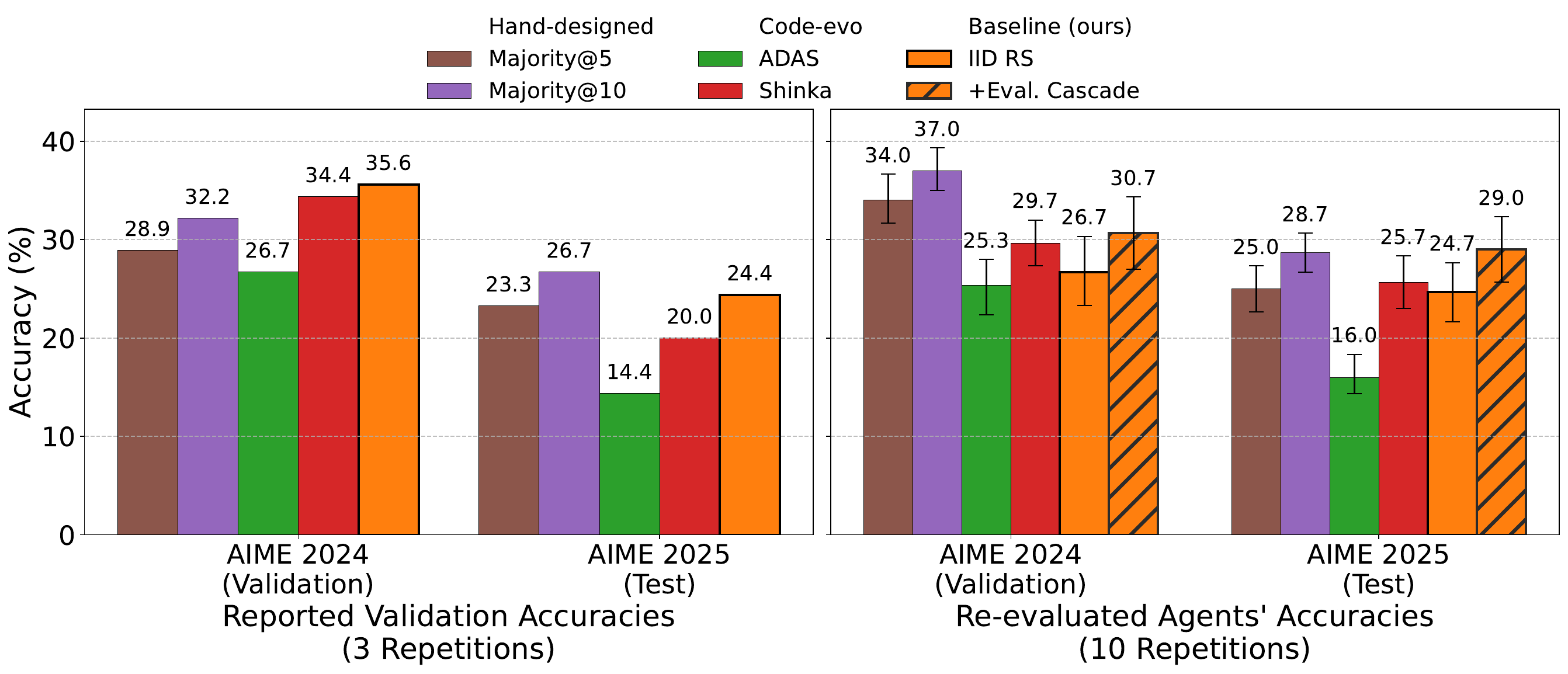}
    \vspace{-3pt}
    \caption{AIME 2024 and 2025 accuracies for different methods. Majority@5/10 indicate manually designed majority vote scaffolds. \textbf{(Left)} Validation accuracies are those measured while evolving different scaffolds, where both validation and test accuracies are based on 3 evaluations over the dataset. ShinkaEvolve numbers are from \citet{lange2025shinkaevolve}. IID RS seemingly performs best out of the search methods, although all do worse than majority vote on the test set, with large drops in accuracy.
    \textbf{(Right)} Results when re-evaluating the scaffolds 10 times, with whiskers denoting 95\% confidence intervals. Validation accuracies are lower for all automated search methods as their scaffolds are seemingly selected more so due to stochasticity in the evaluations than them achieving good performance. Unlike when evaluating only 3 times, it is apparent that there is no clear difference between ShinkaEvolve and IID RS, with the probability of improvement being $P(\text{Shinka}>\text{IID RS})=0.49$. Using an evaluation cascade with IID RS results in selecting a robustly higher-performing scaffold.
    }
    \label{fig:aime_res}
\end{figure*}

To evaluate how baselines compare to code evolution on a problem where the main constraint is the number of function evaluations, we test how well different methods can design agentic scaffolds. In this setting, a method must design a series of LLM calls, called a scaffold, that solves some problem, e.g. finding the correct answer to a math question. A meta-agent (the code evolution pipeline) designs a series of scaffolds, tests how well they do on some validation set, and then picks the best one and evaluates its performance on a test set \citep{hu2024automated, lange2025shinkaevolve}. Typically, the limiting factor is not meta-agent queries but the number of evaluated scaffolds, as depending on which LLM is used each evaluation typically costs \$1-10. For example, \citet{hu2024automated} use training runs of 30 generations, in which we estimate about 50 scaffolds are evaluated, to cost them \$300.

We compare two code evolution methods, ADAS \citep{hu2024automated} and ShinkaEvolve, to IID RS. ADAS runs for 30 generations but can re-evaluate a scaffold to debug it, resulting in up to 90 evaluations. ShinkaEvolve runs for 75 but only about 60 generations result in an evaluated scaffold. Thus, we limit IID RS to evaluating 50 scaffolds, so a scaffold that does not compile does not count towards this limit. We test all methods' ability to design scaffolds for questions from the AIME math competition \citep{AIME2024,AIME2025}, with all agents using GPT4.1-nano for their scaffold evaluations. Note that GPT4.1-nano's knowledge cutoff is June 2024 \citep{openai_gpt41_2025}, before AIME 2025 was released. We compare the discovered scaffolds to a manually designed majority vote@$k$ scaffold. Majority vote means asking the model to answer the question $k$ times and then picking the most prevalent answer, with ties being resolved uniformly at random. Additional technical details are in Appendix \ref{app:agent_evo_details}.

\textbf{Results.} Figure \ref{fig:aime_res} (left) compares each method's scaffolds when evaluating them as done in some prior work, where validation accuracies used throughout the search are reported, with reported test accuracies being over similarly sized datasets. All code evolution methods, including the IID RS baseline, exhibit \textgreater10\% drops in accuracy between their validation and test sets. Although it is likely that AIME 2025 is harder than 2024, this drop is only for automated search methods, with the majority vote baselines degrading by only about 5\%.

\begin{wrapfigure}{r}{0.5\textwidth}
    \centering
    \vspace{-15pt}
    \includegraphics[width=\linewidth]{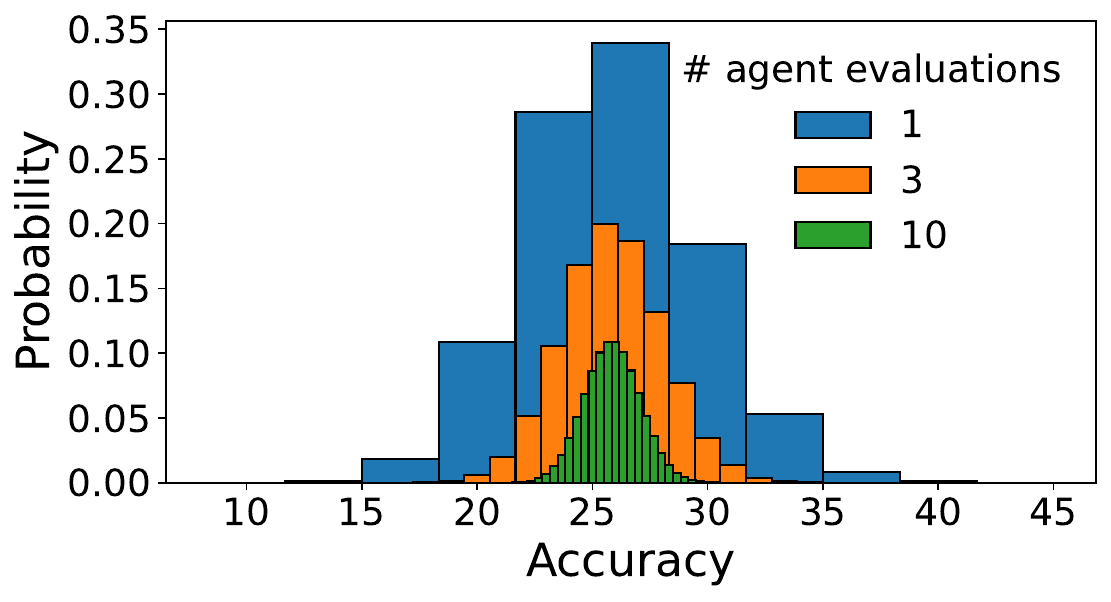}
    \vspace{-17pt}
    \caption{Empirical distributions of majority vote@5 accuracies for AIME 2025 when evaluating each question a different number of times. Even when evaluating 10 times there is a standard deviation of more than 1\% in the accuracy. Distributions were calculated by sampling 100 answers to each question and bootstrapping.
    }
    \label{fig:majv5_dist}
    \vspace{-17pt}
\end{wrapfigure}

Note that to keep the already costly evaluation economical, validation sets used to find scaffolds are typically constructed of $\sim$100 questions (including repetitions), regardless of the full dataset's size \citep[see][Appendix E]{hu2024automated}. However, small datasets introduce high variance into the evaluations. To illustrate, Figure \ref{fig:majv5_dist} shows the empirical AIME 2025 accuracy distribution when using a majority vote@5 scaffold, comparing the uncertainty when each question is evaluated a different number of times. Due to high variance, if scaffolds are evaluated on small datasets then they will be picked not due to their performance but mostly due to the evaluation's stochasticity, resulting in good validation performance being likely coincidental. Figure \ref{fig:aime_res} (right) demonstrates this, as re-evaluating scaffolds discovered by automated methods 10 times shows that their underlying validation accuracies are lower than initially reported.

\vspace{-8pt}
\subsection{Stochasticity Reduction}
\vspace{-6pt}
\label{subsec:stoch_red}
Reducing evaluation stochasticity by na\"ively sampling more would be expensive. Taking AIME 2025 as a typical example, a standard deviation of 0.5\% in the accuracy would require each question to be evaluated roughly $>60$ times, which is $>20\times$ as costly as the default evaluations used by ShinkaEvolve and ADAS. If only a scaffold's final performance is of interest then fewer samples are likely sufficient. Specifically, for reporting final results, we recommend methods to be evaluated over an effective test set size of at least $300$ questions, equivalent to 10 repetitions here, and use 95\% confidence intervals. Higher variance datasets may require more samples.

To reduce costs while also evolving scaffolds, we recommend using an \textit{evaluation cascade}. As low performing scaffolds are not of interest, in a two-level cascade bad scaffolds can be evaluated over a small effective validation set while those that are potentially competitive are then re-evaluated over a larger set to see if their improvement holds. For IID RS this can easily be implemented by re-evaluating the top-X\% of scaffolds using more repetitions. If the top-20\% of scaffolds are re-evaluated then the evaluation cascade is estimated to be $2\times$ cheaper relative to na\"ively using a larger validation set.

An additional source of noise in the evaluations is due to relying on summary statistics instead of comparing distributions. To illustrate, imagine a consistent and noisy scaffold, where the consistent one always gets an accuracy of 20\% while the noisy scaffold has a 90\% probability of getting an accuracy of 12\% but 10\% of the time an accuracy of 100\%. Although in most cases the consistent scaffold outperforms the noisy one, the noisy scaffold has a higher mean accuracy.

To robustly compare two methods' test accuracies, we recommend using the \textit{probability of improvement}, a metric for comparing two distributions. Introduced by \citet{agarwal2021deep} for comparing deep reinforcement learning methods, the probability of improvement measures the probability a single run of $A$ would outperform a single run of $B$, $P(A>B)$.

However, when evolving scaffolds or picking the best one, the comparisons are between sets of scaffolds and not pairs. To robustly compare sets of scaffolds, we introduce the \textit{probability of dominance}, a generalization of the probability of improvement that tests whether a method is better than a set of alternatives. Explicitly, given methods (here, scaffolds) $A_1,A_2,...,A_M$, the probability of dominance is the empirical estimate of $P(A_1>A_2,A_3,...,A_M)$, which is the probability a given run of $A_1$ would outperform (``dominate'') a single run of each of $A_2,A_3,...,A_M$. On both the probability of improvement and dominance see Appendix \ref{app:prob_imp} for an extended discussion.

Using the probability of dominance with an evaluation cascade allows IID RS to find a better scaffold that robustly generalizes to the test set, see right plot in Figure \ref{fig:aime_res}. This scaffold outperforms all scaffolds found without robust evaluations and performs comparably well to majority vote@10.

\vspace{-6pt}
\section{Machine Learning Competitions}
\vspace{-4pt}
To see how the baselines compare to code evolution for tasks with limited wall-clock time, we test their performance on MLE bench, a benchmark of Kaggle competitions \citep{chan2024mle}. For each competition, each method is given 24 hours to produce a high-performing solution. All methods select their best-performing solution based on its validation accuracy. All runs are given a single RTX 8000 GPU with 12 CPU cores. We use the ten-competition subset of \citet{schmidgall2025agent} for evaluating the different methods.

We compare the baselines to AIDE, a code evolution pipeline designed specifically for Kaggle competitions \citep{jiang2025aide}. \citet{chan2024mle} find that on MLE bench AIDE outperforms other code evolution systems, such as MLAB \citep{huang2023mlagentbench} and OpenHands \citep{wang2024openhands}. Additional technical details are included in Appendix \ref{app:mle_bench_details}.

\textbf{Results.} Table \ref{tab:mle_bench} shows each method's results for each competition. SCS and AIDE perform similarly well, with SCS beating it on 6/10 competitions. Simple baselines can thus perform well also in difficult time limited domains.

\begin{table*}[!h]
    \centering
    \vspace{-4pt}
    \caption{Results for different methods on the Kaggle competitions. ${\color{brown}*},{\color{gray}+},{\color{amber}\times}$ denote scores that would have gotten bronze, silver, or gold medals respectively. Best and second-best method results per competition are respectively shown in bold and underlined. For unabbreviated competition names and human baseline metrics, see Appendix \ref{app:mle_bench_details}.}
    \label{tab:mle_bench}
    \begin{tabular}{l|ccc}
        Competition 
        & AIDE 
        & IID RS 
        & SCS \\
        \midrule
        \# competitions $\geq$ AIDE & & 2/10 & 6/10 \\
        \midrule
        \# competitions $\geq$ Median
        & 8 & 4 & 8 \\
        \# competitions $\geq$ Bronze
        & 5 & 3 & 5 \\
        \# competitions $\geq$ Silver
        & 3 & 3 & 5 \\
        \# competitions $\geq$ Gold
        & 2 & 2 & 2 \\
        \midrule
        Insult Detection \arrowtag{\uparrow}
        & \textbf{0.9135}$^{\color{amber}\times}$ 
        & \underline{0.9056}$^{\color{amber}\times}$ 
        & 0.8460$^{\color{amber}\times}$ \\

        Dec. 2021 Tabular \arrowtag{\uparrow}
        & 0.9620$^{\color{amber}\times}$ 
        & \underline{0.9621}$^{\color{amber}\times}$ 
        & \textbf{0.9623}$^{\color{amber}\times}$ \\

        Trans. Conductors \arrowtag{\downarrow}
        & \underline{0.0619}$^{\color{gray}+}$ 
        & 0.0620$^{\color{gray}+}$ 
        & \textbf{0.0616}$^{\color{gray}+}$ \\

        English Text Norm. \arrowtag{\uparrow}
        & \underline{0.9823} 
        & 0.9332 
        & \textbf{0.9913}$^{\color{gray}+}$ \\

        May 2022 Tabular \arrowtag{\uparrow}
        & \textbf{0.9965} 
        & 0.9140 
        & \underline{0.9839} \\

        Random Pizza \arrowtag{\uparrow}
        & \underline{0.6820} 
        & 0.6206 
        & \textbf{0.6891} \\

        Spooky Author \arrowtag{\downarrow}
        & \textbf{0.2883}$^{\color{brown}*}$ 
        & 0.4468 
        & \underline{0.4031} \\

        Toxic Jigsaw \arrowtag{\uparrow}
        & \underline{0.9807} 
        & 0.9780 
        & \textbf{0.9863}$^{\color{gray}+}$ \\

        Russ. Text Norm. \arrowtag{\uparrow}
        & \textbf{0.9756}$^{\color{brown}*}$ 
        & \underline{0.9735} 
        & 0.9453 \\

        NYC Taxi \arrowtag{\downarrow}
        & 11.460 
        & \textbf{5.1483} 
        & \underline{5.9758} \\        
    \end{tabular}
    \vspace{-18pt}
\end{table*}

\section{Combinatorial Optimization}
ALE-Bench \citep{imajuku2026ale} tests how well AI systems can design heuristics for difficult optimization problems, many likely being NP-hard. As the agentic scaffold evaluations proved highly stochastic, we use ALE-Bench LITE to test how the two baselines compare to complicated code evolution pipelines.

We benchmark different methods on the 10 ALE-Bench LITE tasks, generally following the setup used by \citet{lange2025shinkaevolve}. Notably, instead of a special AWS instance we run problems locally and adjust per-problem timeouts accordingly so the initial program runs successfully. All methods are run on identical machines, allowing comparisons between them, but making comparisons with methods benchmarked using a different setup not be apples-to-apples. We report the performance of the best program per problem per method on the public test set, which corresponds to the top-1 evaluation of \citet{lange2025shinkaevolve}. This evaluation has a low variance, so there is no need for an evaluation cascade or other more robust metrics.

All methods  -- ShinkaEvolve, IID RS, and SCS -- generate 50 programs per problem. SCS has a single trial and 10 programs per generation. ShinkaEvolve uses the default hyperparameters used for the ALE-Bench experiments in \citet{lange2025shinkaevolve}, except with the non-Gemini models being disabled and using Gemini-2.5 Flash instead of GPT-5 mini as the meta-model.

\begin{figure}[!h]
    \centering
    \includegraphics[width=\linewidth]{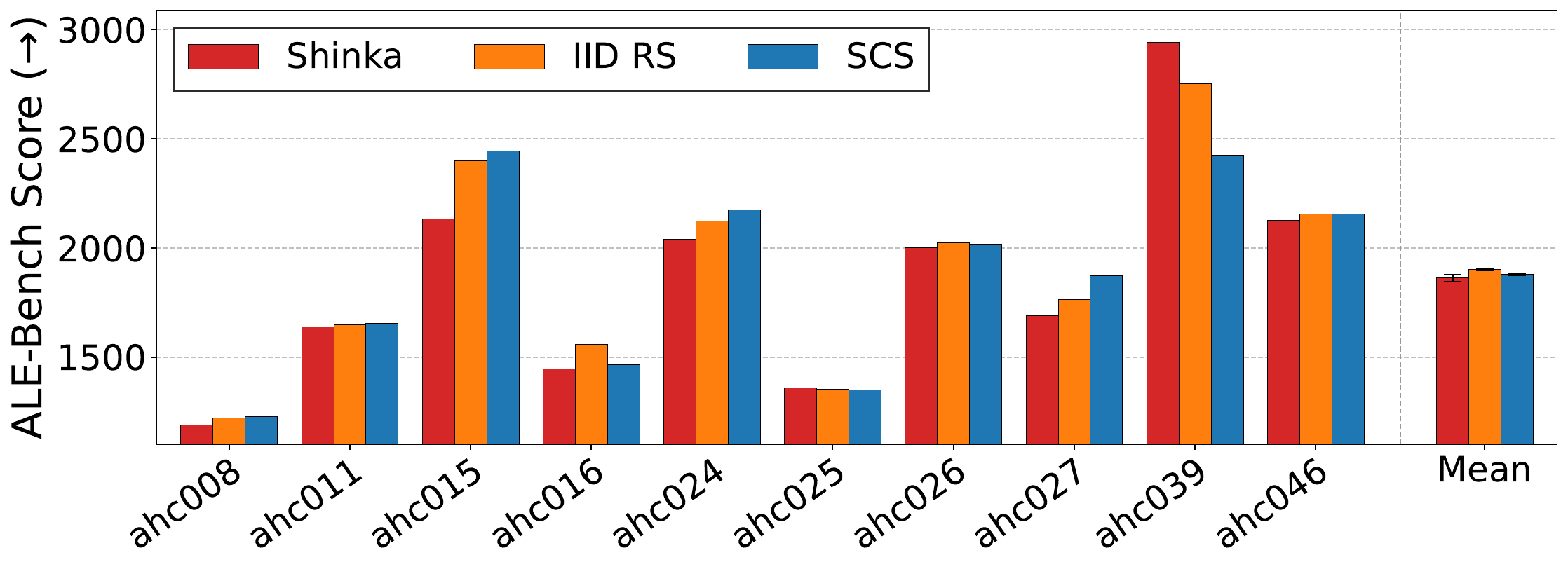}
    \caption{Relative performance of different methods on ALE-Bench LITE. IID RS and SCS outperform ShinkaEvolve on 8/10 problems and have a higher average score. Scores are the median of three evaluations of the best discovered program. Means are over the three evaluations, with uncertainties denoting standard errors.}
    \label{fig:ale}
\end{figure}

Figure \ref{fig:ale} shows the results. Both IID RS and SCS beat Shinka on 8/10 problems, with SCS' average performance being worse chiefly due to poor performance on problem ahc039, which may be an outlier. 

Initially, we were surprised to see that both methods outperform ShinkaEvolve so clearly. There are a few possible reasons why this may be, one being that Shinka is inefficient given a finite sample budget due to sometimes doing file diffs instead of full file rewrites. Another theory is that ShinkaEvolve’s strictly sequential structure can bias it down potentially suboptimal search paths, while both methods -- especially IID RS -- are relatively less biased. Whichever theory is correct, note that both support a similar point -- that most design choices in existing complicated pipelines are likely redundant, if not detrimental.

\section{Discussion}
\vspace{-5pt}
\label{sec:discuss}
Across several domains, simple baselines are competitive with domain-specific sophisticated code evolution pipelines. This remains true across different constraints: a limited API budget when finding mathematical bounds, a limit on the number of function evaluations when designing agentic scaffolds or heuristics for combinatorial optimization, and limited wall-clock time in machine learning competitions. Accordingly, the main drivers for \textit{improving} performance further come not from adding complexity to the search pipeline but parts of the setup external to the search pipeline, such as the search space, the domain knowledge, or the evaluation framework. Comparisons where part of the setup beyond the pipeline differs can lead to brittle conclusions on a method's performance.

These results likely hold also for much larger budgets. All budgets used here are either standard or above-standard for the relevant settings, e.g. with ShinkaEvolve using only \$12 and not \$20 for their circle packing experiments \citep{lange2025shinkaevolve}. Additionally, after this paper was released, \citet{tsoukalas2026advancing} noted that a very simple code evolution pipeline was competitive with AlphaEvolve in finding proofs for open mathematical theorems. \citet{tsoukalas2026advancing} find similar results when using hundreds of dollars per experiment, beyond the tens of dollars used here.

Code evolution has achieved impressive capabilities across a range of domains, yet which design choices contribute to these successes has been largely overlooked. We find that much of the additional complexity in existing sophisticated pipelines does not translate to further improvements in performance. \textit{Select} design choices may still be important for a specific domain but not others, see the need for textual feedback for MLE bench (Appendix \ref{app:mle_bench_details}).

Why were the discovered shortcomings not identified previously? First, for many works in code evolution, different contributions may be conflated, leading baselines to often be ignored. For example, \citet{novikov2025alphaevolve} give two main contributions: they both introduce a search method, AlphaEvolve, and demonstrate various scientific discoveries and optimizations it achieved. Both are likely valuable in and of themselves, but achieving a scientific discovery does not necessarily imply that the method used to find it is performant. 

Second, if not tightly controlling for fair comparisons, there can be apparent improvements which in practice stem not from the code evolution pipeline but something else, such as domain knowledge. Such accidental improvements can easily happen, as different published results are often used without rerunning exact setups, with matters like prompts potentially being seen as unimportant.

To facilitate better progress in the field, we recommend the following best practices for future work. \textbf{If proposing search methods, run fair comparisons and compare to simple baselines.} Comparisons should use the same language models, domain knowledge, search spaces, and so on. These are important not only for sharing robust results but for developing methods, as otherwise apparent improvements may not generalize across settings. The simple baselines and setup here provide a convenient test bed for when discrete changes to a pipeline are beneficial. \textbf{If showcasing a scientific discovery, disclose the domain knowledge and search setup used to find it.} The domain knowledge in prompts or initial programs, coupled with the search's structure, may take a larger part in enabling a discovery than the pipeline itself. Without disclosing exact setups the discovery might not be reproducible, nor allow being built upon. When evaluating noisy functions, such as when searching for agentic scaffolds, \textbf{use evaluation cascades and the probability of dominance to economically get robust results that better generalize.} Otherwise, scaffolds may underperform relative to simple existing setups. In general, the selected noisy discovered functions should be re-evaluated several times and with reported confidence intervals, to ensure reported results are robust. 

To better enable rigorous code evolution development we include in Appendix \ref{app:cheap_code_evo} a set of tips to keep the development economical, allowing more runs to be used and different setups compared. Altogether, we hope these recommendations and findings will enable continued development of code evolution that matters.

\section*{Acknowledgements}
Thanks to much of the Sakana AI research team for fruitful discussions, and especially to Robert Lange for helping with the ShinkaEvolve runs and discussing the work throughout. Thanks as well to Yujin Tang, who provided good feedback and support that spurred the development of an earlier version of this work. Thanks also to Michael Beukman, Dulhan Jayalath, and Anushka Nair for feedback on an earlier draft, to Noya Gideoni and Katrina Dickson for proofreading it, and to Edan Toledo for feedback and some help with the MLE bench setup.

We thank the Oxford ARC cluster for providing the GPUs for the MLE bench experiments. Yonatan is funded by the Rhodes Trust and the AIMS EPSRC CDT (grant no. EP/S024050/1).

\bibliography{bib}
\bibliographystyle{plainnat}

\appendix

\newpage

\section{Related work}
\label{app:relwork}
While modern code evolution relies on LLMs, code evolution as a term has existed at least as far back as 1999 \citep{keller1999evolution}. Although only an emerging field, there are a plethora of works on code evolution, we focus here on some notable representatives. Some earlier LLM-based code evolution methods have components similar to the two given baselines. FunSearch \citep{romera2024mathematical} and AlphaCode \citep{li2022competition} have components similar to IID RS.

There are also works that implement pipelines similar to SCS, often with domain-specific adjustments. For example, \citet{ma2023eureka} use an SCS-like code evolution pipeline to find better reward functions for reinforcement learning environments.

In other subfields, especially reinforcement learning, there have been many efforts to improve the field's robustness. \citet{henderson2018deep} discuss difficulties in reproducing existing work in RL while \citet{agarwal2021deep} showcase problems in evaluations, proposing various best practices. \citet{goldie2025should} compare a variety of meta-RL algorithms, discussing best practices for future work.

Regarding agentic scaffolds, \citet{el2025inefficiencies} discuss inefficiencies in searching for agentic scaffolds. They note that typical scaffolds are expensive to run when measured based on the price per correct answer, but do not compare to different search methods or propose methods to make the search cheaper.

\section{AlphaEvolve Problems}
\label{app:probs}
Table \ref{tab:probs} lists the 9 problems used in §\ref{sec:mathbounds}, taken from \citet{novikov2025alphaevolve}.

\begin{table}[h]
    \centering
    \caption{Bounds of AlphaEvolve (AE) problems studied here, divided into analysis (top), combinatorics (middle), and geometry (bottom). The arrow next to the problem name indicates whether it is an upper bound, so lower results are tighter and hence better ($\downarrow$), or a lower bound so higher is tighter and thus better ($\uparrow$). All numbers are from \citet{novikov2025alphaevolve}. ``Pre-AE'' are the best bounds from before \citet{novikov2025alphaevolve}.}
    \vspace{2pt}
    \begin{tabularx}{\textwidth}{@{}>{\raggedright\arraybackslash}p{2.9cm}>{\raggedright\arraybackslash}X@{}ccc@{}}
        Problem & Input size & Pre-AE Bound & AE Bound & AE Appendix \\
        \midrule
        First autocorrelation inequality ($\downarrow$) & Unbounded, step function heights & 1.5098 & 1.5053 & B.1 \\
        \addlinespace
        Second autocorrelation inequality ($\uparrow$) & Unbounded, step function heights & 0.88922 & 0.8962 & B.2 \\
        \addlinespace
        Uncertainty inequality ($\downarrow$) & 3 coefficients of a Hermite polynomial & 0.3523 & 0.3521 & B.4 \\
        \addlinespace
        \cmidrule(lr){1-5}
        Erdős' minimum overlap ($\downarrow$) & Unbounded, step function heights & 0.380927 & 0.380924 & B.5 \\
        \addlinespace
        Sums vs. differences of finite sets ($\uparrow$) & Unbounded, set $U\subset\mathbb{Z}_{\ge 0}$ fulfilling some properties & 1.14465 & 1.1584 & B.6 \\
        \addlinespace
        \cmidrule(lr){1-5}
        Max–min distance ratio for 16 2D points ($\downarrow$) & 32 coordinates (16×2) & 12.890 & 12.88927 & B.8 \\
        \addlinespace
        Heilbronn triangles $n=11$ ($\uparrow$) & 22 coordinates (11×2) in a unit-area triangle & 0.036 & 0.0365 & B.9 \\
        \addlinespace
        Kissing number in 11D ($\uparrow$) & Largest number of 11D sphere centers all tangent to a common sphere & 592 & 593 & B.11 \\
        \addlinespace
        Circle packing ($\uparrow$) & 78 -- 26 center coordinates (26×2) and 26 radii & 2.634 & 2.63586 & B.12 \\
        \addlinespace
    \end{tabularx}
    \label{tab:probs}
\end{table}

\section{Finding Bounds for Math Problems: Technical Details}
\label{app:math_techdet}
For the baselines we use Gemini 2.5 Pro, sampling with a temperature of $0.8$, a top-$p$ sampling cutoff of $0.95$, a thinking budget of 1024 tokens, and let each program evaluation run for at most 5 minutes. These settings were not thoroughly tuned and were chosen as they seemed like sensible defaults. IID RS samples 2000 programs and SCS samples 20 programs per generation, having 10 generations per trial and 6 trials overall, resulting in 1200 generated programs. This is where the numbers in Figures \ref{fig:iid_rs} and \ref{fig:scs} come from. After the first generation in each trial, SCS randomly picks 3 programs that successfully ran in the previous generation and appends them and their bounds to the IID RS prompt when generating a new program.

For fairness, ShinkaEvolve had some of its hyperparameters slightly tuned, as at its default settings it was not competitive with the baselines. Most importantly, it also uses a thinking budget of 1024 tokens, as otherwise it is much more expensive and results in far fewer generations. Other hyperparameters are taken from ShinkaEvolve's default circle packing configuration, except for using 5 island subpopulations instead of 2 due to the slightly larger budget of \$20 instead of the default setup's \$12, and uses 2 programs from its archive for inspiration instead of the default 4 in order to reduce API costs.

\section{OpenEvolve Results on Math Problems}
\label{app:openevolve}
When running OpenEvolve \citep{openevolve} using Gemini models we found it to often crash due to getting \texttt{None} results, while also not having a way to specify LLM thinking budgets or a max API budget for a run. Table \ref{tab:openevolve_res} shows its results given 300 generations when using GPT-5.2 and GPT-5 mini for the problems it managed to run successfully. For the six problems it ran successfully, OpenEvolve got an average subpar rank of 3.08 relative to the baselines and ShinkaEvolve.

\begin{table*}[!h]
    \caption{Best bounds found by OpenEvolve given 300 generations using GPT-5.2 and GPT-5 mini. ``--'' denotes a failed run. Arrows denote whether higher or lower is better.}
    \centering
    \begin{tabular}{L{6.5cm}c}
        Problem & OpenEvolve \\
        \midrule
        First autocorr. ineq. \arrowtag{\downarrow} & -- \\
        Second autocorr. ineq. \arrowtag{\uparrow} & 0.8903 \\
        Uncertainty ineq. \arrowtag{\downarrow} & 0.3521 \\
        Erd\H{o}s' min. overlap \arrowtag{\downarrow} & -- \\
        Sums/differences of sets \arrowtag{\uparrow} & 1.1095 \\
        Max--min dist. ratio \arrowtag{\downarrow} & -- \\
        Heilbronn triangles \arrowtag{\uparrow} & 0.0354 \\
        Kissing number in 11D \arrowtag{\uparrow} & 348 \\
        Circle packing \arrowtag{\uparrow} & 2.541 \\
    \end{tabular}
    \label{tab:openevolve_res}
\end{table*}

\section{Minimal Domain-Knowledge Circle Packing Prompt}
\label{app:min_cp_prompt}
The \verb|verify_circles| function is taken from AlphaEvolve's validation script. \verb|${max_execution_time}| is replaced with the time limit per problem, which in practice was 300 seconds (5 minutes) for the programs in Table \ref{tab:results_math}. The minimal domain knowledge prompt used for all tested methods in Table \ref{tab:results_math} is:
\begin{minted}[breaklines,
autogobble,
fontsize=\scriptsize,
frame=single]{markdown}
You are an expert programmer specialising in numerical optimisation. Implement a Python function with the exact signature:

def pack_circles() -> Tuple[np.ndarray, np.ndarray, float]:

The function must pack 26 non-overlapping circles into the unit square [0,1]x[0,1] so that the sum of their radii is maximised. Returns:
- centers: np.ndarray of shape (26, 2) with (x, y) coordinates
- radii:   np.ndarray of shape (26,) with positive radii
- sum_radii: float = radii.sum()

You can use these predefined helper functions without redefining them:
```
import numpy as np
import itertools

def verify_circles(circles: np.ndarray) -> bool:
    """Checks that the circles are disjoint and lie inside a unit square.

    Args:
      circles: A numpy array of shape (num_circles, 3), where each row is
        of the form (x, y, radius), specifying a circle.

    Returns:
      True if all circles are disjoint and fully inside the unit square,
      False otherwise.
    """
    # Check pairwise disjointness.
    for circle1, circle2 in itertools.combinations(circles, 2):
        center_distance = np.sqrt((circle1[0] - circle2[0])**2 + (circle1[1] - circle2[1])**2)
        radii_sum = circle1[2] + circle2[2]
        if center_distance < radii_sum:  # Overlap
            return False

    # Check all circles lie inside the unit square [0,1]x[0,1].
    for circle in circles:
        x, y, r = circle
        if x - r < 0 or y - r < 0 or x + r > 1 or y + r > 1:
            return False

    return True
```

All circles must be fully inside the square and not overlap. You have up to ${max_execution_time} seconds for your solution to run. Please only supply the code for pack_circles, please define helper functions inside it.
\end{minted}

\section{Domain Knowledge Ablations}
\label{app:dk_ablate}
To see whether simple methods can leverage domain knowledge similarly to complicated code evolution pipelines, and whether prompts with domain knowledge transfer between methods, we run several ablations. First, in the workshop version of this work we found that giving IID RS a better prompt and 1000 samples led it to find the same ``optimal'' circle packing bound that ShinkaEvolve finds. Specifically, the prompt included a compute max radii function akin to the one that ShinkaEvolve has by default, but one which is exact and not a heuristic. Thus, simple methods can leverage domain knowledge as well.

Note that ShinkaEvolve's expert prompt was designed specifically for ShinkaEvolve, and likely specifically for the kind of errors it makes. We speculate it was designed by running ShinkaEvolve, seeing which errors it makes, and then adding instructions in the prompt to mitigate those errors. This is further supported by seeing that it finds the same “optimal” circle packing only 1/3 times when not given this domain knowledge, see section \ref{subsec:dk}. Thus, if method 1 makes type A errors 50\% of the time and type B errors the other 50\%, while method 2 makes type A errors 30\%, type B 20\%, and type C 50\%, a prompt saying ``don’t do A'' would help method 1 much more than method 2, and for method 2 might even be detrimental. 

This is what we observe in practice. To test whether ShinkaEvolve's prompt transfers to other methods, we test how well the two baselines perform given ShinkaEvolve's circle packing prompt, see Listing \ref{lst:shinka_cp_prompt}. Given ShinkaEvolve's ``expert'' prompt, IID RS and SCS find circle packings with a sum of radii of 2.626 and 2.621 respectively, both worse than when they are given no domain knowledge at all. These results imply domain knowledge, depending on how it is used, can be detrimental to one method while being beneficial to another.

\section{Bootstrap Estimate}
\label{app:bootstrap}
\textbf{IID Random Sampling.} Here the bootstrap estimate amounts to calculating a pass@k. Under a given budget we find the last generation ShinkaEvolve reached before exceeding it, and as its score take the best bound found until then. If the budget allows sampling $k$ IID programs then the probability of sampling one that matches or outperforms Shinka's score can be approximated by repeatedly sampling $k$ programs out of the $n$ total generated, here being 2000, and seeing if any match or exceed Shinka's score. If $c$ is the number of programs that match/exceed code evolution's score this probability estimate can be analytically calculated as $1 - \frac{{n-c \choose k}}{{n \choose k}}$, as per \citet{chen2021evaluating}. The average IID cost per program is estimated as the average cost over all sampled programs, so small cost differences between individual generations are ignored.

\textbf{Sequential Conditioned Sampling.} For the SCS bootstrap estimate it is important to consider a realistic scheme of how a given budget would be exhausted, with different setups potentially yielding different results. Here it is assumed that a budget is used up, generation by generation, until a trial is exhausted, after which a new trial is begun. The bootstrap is estimated by picking a random trial, seeing up until which generation in it can be searched under the given budget, and if the trial is exhausted then continuing to one of the remaining trials. This is how the results in Table \ref{tab:results_math} were picked, except after exhausting a trial the next was picked not at random but chronologically.

Formally, if $p_{\geq\text{Shinka}}(B,T)$ is the probability of matching/exceeding a bound $s$ given a budget $B$ and a set of trials $T$, where trial $t \in T$'s budget is $b_t$, then $p_{\geq\text{Shinka}}$ is recursively defined as

\begin{equation}
    p_{\geq\text{Shinka}}(B,T) = \frac{1}{|T|} \sum_{t \in T} \max(\mathbbm{1}_{\max(t) \geq s}, p_{\geq\text{Shinka}}(B - b_t, T \setminus \{t\}))
\end{equation}

This assumes that all $T$ trials are within the budget, where in practice some might not be while others are. In these cases the sum is only over the trials within budget. For minimization problems the $\max(t)\geq$ should be changed to $\min(t)\leq$.

\begin{figure}[!h]
    \centering
    \includegraphics[width=0.8\linewidth]{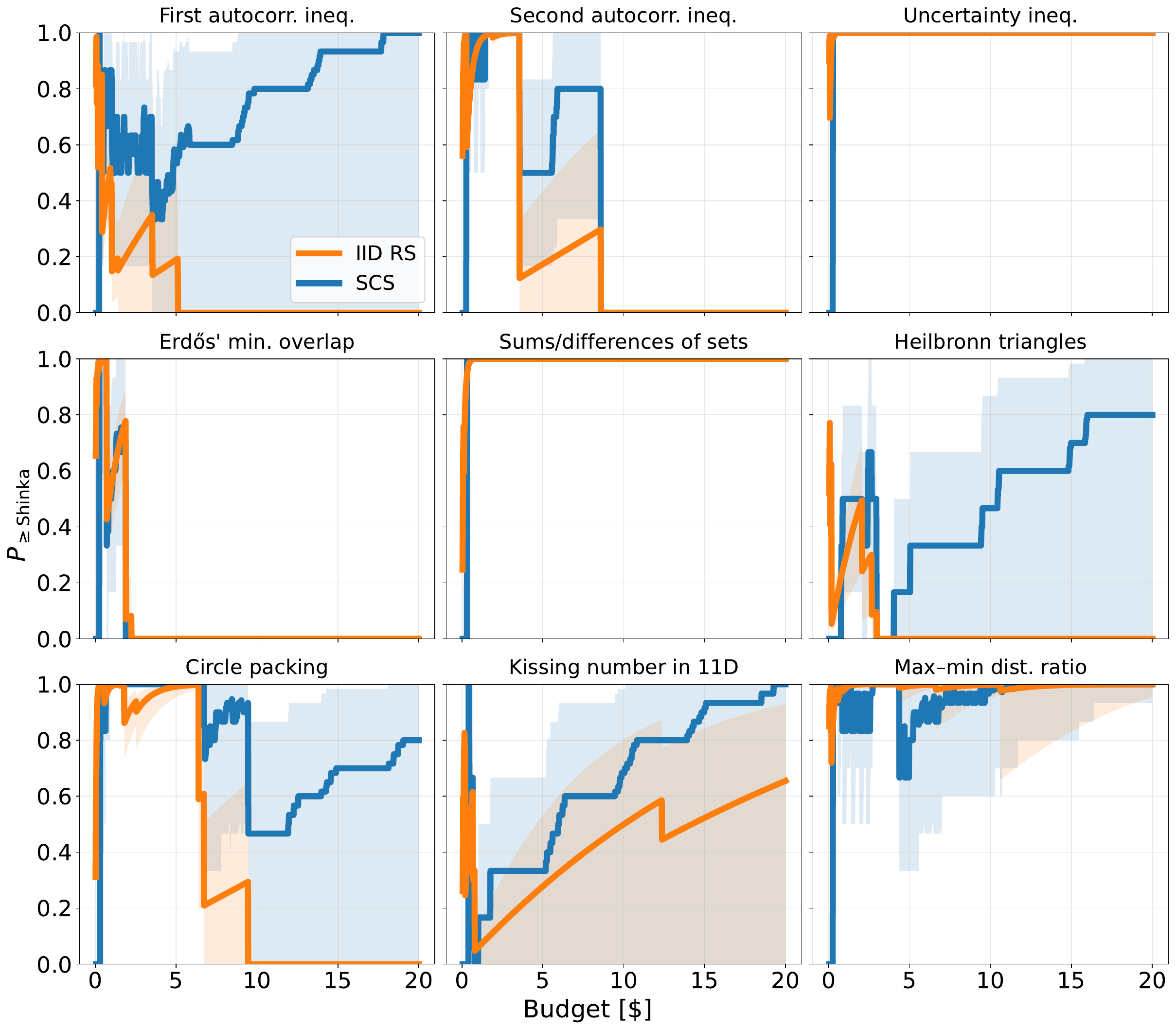}
    \caption{Per-problem probability of matching/exceeding ShinkaEvolve for the two baselines. Shaded regions are 95\% confidence intervals.}
    \label{fig:prob_imp_per_prob}
\end{figure}

\textbf{Confidence Intervals.} In both cases confidence intervals are calculated by bootstrapping over the bootstrap, i.e. picking programs/trials with replacement, calculating their probabilities of matching/exceeding, and then calculating the 2.5\% and 97.5\% quantiles over this empirical distribution over the matching/exceeding probabilities. Note that the uncertainties, e.g. in Figure \ref{fig:prob_imp_per_prob}, are large as there is often a >2.5\% probability of getting a worse answer than ShinkaEvolve. However, these estimates are likely overly negative as there exists counterfactual information for worse answers -- other sampled programs -- but not for getting a better answer. This coupled with the probability of matching/exceeding's calculation being nonlinear is why the uncertainties are asymmetric and tend to have low lower bounds.

\section{Figure \ref{fig:prob_imp} Per-problem Breakdown}

\label{app:per_prob_p_imp}

\section{Math Bounds Sample Efficiency}
\label{app:math_sample_eval_effic}
\begin{figure}[!h]
    \centering
    \includegraphics[width=0.85\linewidth]{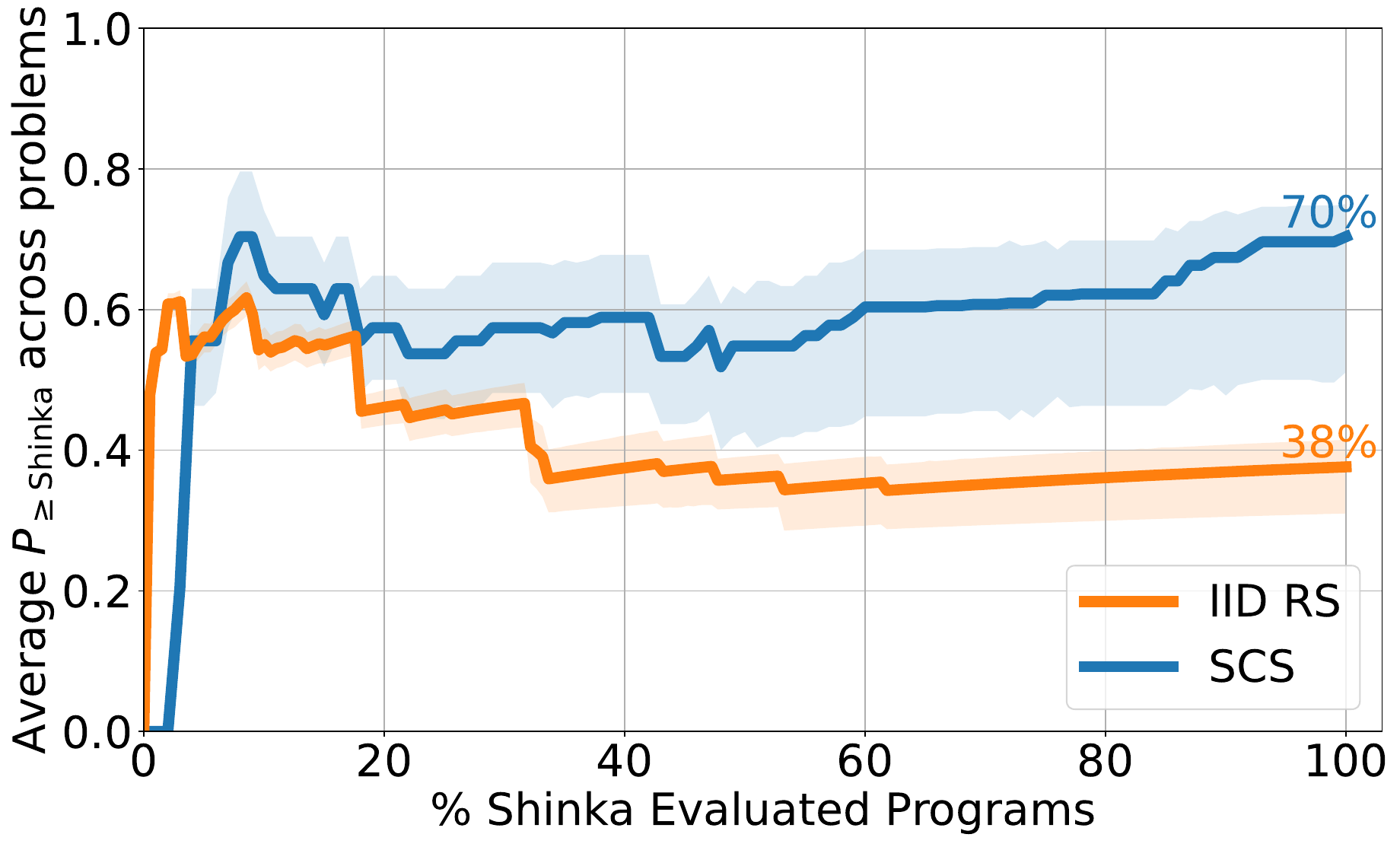}
    \caption{For each baseline, the average probability of matching or exceeding ShinkaEvolve across the 9 problems, as a function of the number of evaluated programs. The $x$ axis is the percent of evaluated programs out of the maximum in the corresponding ShinkaEvolve run. Shaded regions are 95\% confidence intervals.}
    \label{fig:prob_imp_evals}
    \vspace{-10pt}
\end{figure}

\begin{figure}[!h]
    \centering
    \includegraphics[width=0.9\linewidth]{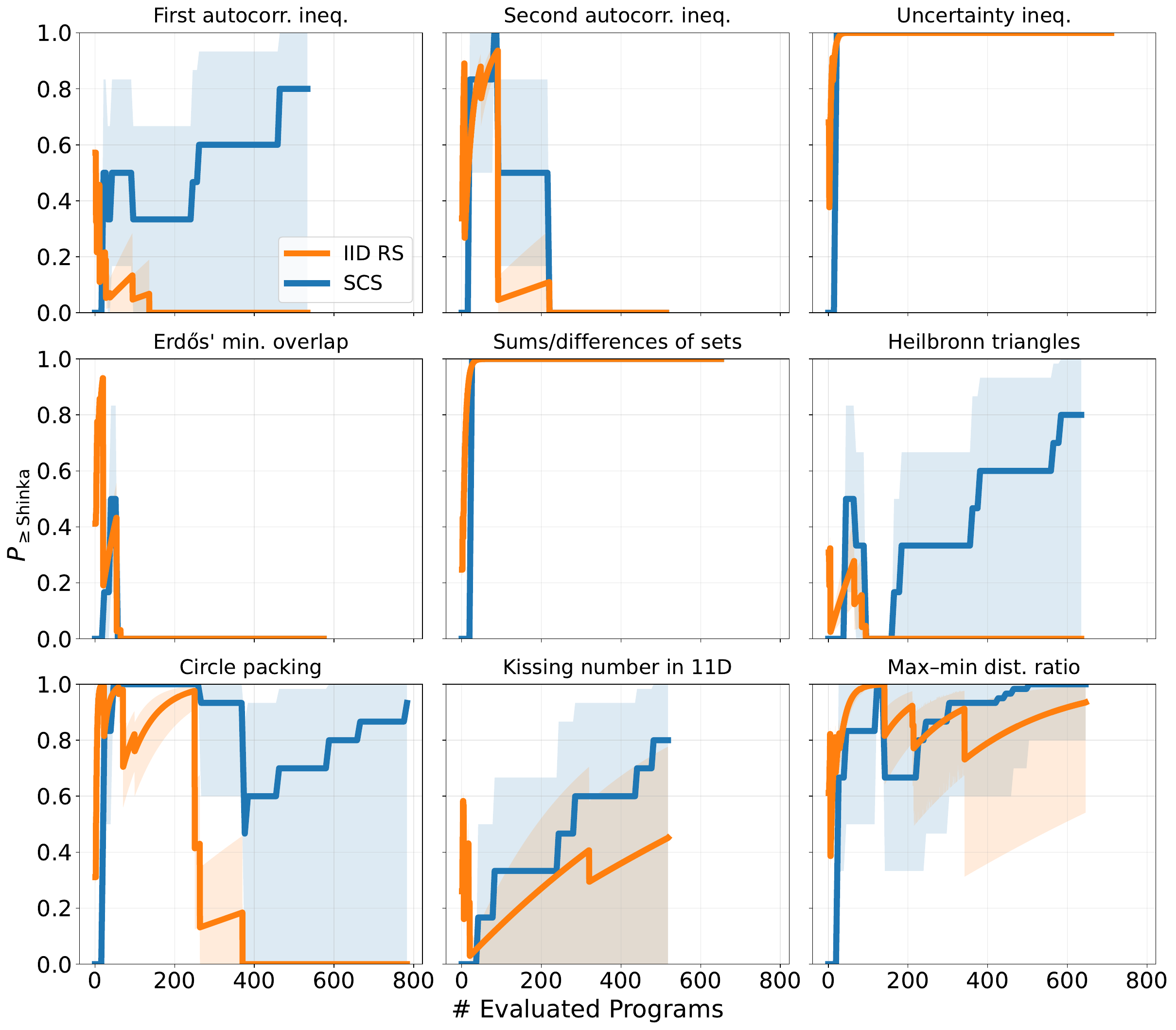}
    \caption{For each baseline, the per-problem probability of matching or exceeding ShinkaEvolve across the 9 problems, as a function of the number of evaluated programs. Shaded regions are 95\% confidence intervals.}
    \label{fig:prob_imp_evals_per_prob}
\end{figure}

Figures \ref{fig:prob_imp_evals} and \ref{fig:prob_imp_evals_per_prob} show respectively the aggregate and per-problem probability of the baselines matching/exceeding Shinka when the budget is defined not by the API cost but the number of evaluated programs. ShinkaEvolve evaluates one program per generation while the baselines evaluate one program per sample. Note that for a set API budget Shinka runs until different numbers of generations, as the average cost per program differs per problem. 

Shinka being a bit cheaper per program, perhaps due to also using some cheaper LLMs, results in the final probabilities in Figure \ref{fig:prob_imp_evals} being lower than those in Figure \ref{fig:prob_imp}, but only slightly.

\section{Different Problem Formulations}
\label{app:prob_search_space}

\subsection{Improved Uncertainty Inequality Formulation}
\label{app:uncert_ineq_form}
We first describe the problem in its generality, based on Appendix B.4 of \citet{novikov2025alphaevolve} and \citet[p.~679]{gonccalves2017hermite}. For a function $f:\mathbb{R}\to\mathbb{R}$ define its Fourier transform as $\hat{f}(x)=\int_{-\infty}^\infty f(t)e^{-2\pi i xt}dt$. Let the radius of the smallest disc for which outside of it $f$ is nonnegative be defined as $A(f)\coloneqq \inf(\{r>0|\forall |x|\geq r:f(x)\geq 0\})$. In the uncertainty inequality problem we wish to find the largest constant $C$ for which $A(f)A(\hat{f})\geq C$, under the conditions where a) $f$ is even and b) $\max(f(0),\hat{f}(0))\leq0$.\footnote{For this last condition \citet{novikov2025alphaevolve} mistakenly state it as $<0$, see \citet[p.~679]{gonccalves2017hermite}.}

Denoting the $n$th Hermite polynomial as $H_n$, \citet[p.~697]{gonccalves2017hermite} show that functions of the form $f(x)=\sum_{n=0}^\infty \alpha_nH_{4n}(\sqrt{2\pi}x)e^{-\pi x^2}$ fulfill the two conditions given that the coefficients $\alpha_n$ are chosen so $f(0)=0$. As here $\hat{f}(x)=f(x)$, this automatically fulfills condition b). As even Hermite functions are even, this fulfills condition a).

\citet{gonccalves2017hermite} construct their upper bound of 0.3523 by setting all $\alpha_n$ except for $\alpha_0,\alpha_1,\alpha_2,\alpha_3$ to zero and numerically finding which $\alpha$s minimize $C$. This is the formulation also used by \citet{novikov2025alphaevolve} and in our Table \ref{tab:results_math}.

We modify this formulation in two ways. First, \citet{novikov2025alphaevolve} use physicist's Hermite polynomials, where the leading coefficient of $H_n$ is $2^n$. This leads to numerical instabilities when attempting to use higher orders. Instead, we use the probabilist's Hermite polynomials, which are the same but rescaled, so $He_n\coloneqq \frac{H_n}{2^n}$, resulting in the leading coefficient for all polynomials being one. Our second modification is setting all $\alpha$s beyond $\alpha_7$ to zero instead of $\alpha_3$. This allows $f(x)$ to represent a larger class of functions. It is likely possible going beyond $\alpha_7$ and reducing the bound further but we encountered numerical instabilities when trying to do so, likely from using very high order polynomials.

\subsection{Different Formulations for Additional Problems}
\label{app:prob_form_extra_probs}
It is simple to illustrate that larger search spaces lead to better performance by showing that existing methods perform worse when a problem's search space is artificially restricted. We demonstrate this over three problems, where in all of them all tested methods perform similarly well.

For circle packing, when all methods are limited to using circles with a maximum radius of 0.1, ShinkaEvolve and the two baselines all find the same bound of 2.514. For reference, the maximum radius found in good solutions for the unconstrained case is about 0.13, with those solutions achieving a bound of about 2.64.

For the second autocorrelation inequality, when the constructed step function is limited to use only 10 steps instead of an arbitrary amount, ShinkaEvolve finds a bound of 0.868, while the baselines find a new bound of 0.867. For reference, the AlphaEvolve bound is $\sim$0.896.

For the Erd\H{o}s' minimum overlap problem, when also restricted to 10 steps all methods find a worse bound of 0.3824, up from the known best of 0.3809.

\section{Agentic Scaffold Evolution Technical Details}
\label{app:agent_evo_details}
\citet{lange2025shinkaevolve} use Gemini-2.5 Pro, GPT-o4 mini, and Claude 4 Sonnet for ShinkaEvolve's meta-agent when designing scaffolds. IID RS here uses their prompt and with Gemini-2.5 Pro, without a limit on the thinking budget as the cost for designing a scaffold is almost negligible relative to the evaluation costs. ADAS is run using GPT-4o and by manually changing the ADAS prompt for the GPQA dataset to fit AIME's format, as ADAS has specialized prompts to fit its general setup. We attempted running ADAS with Gemini-2.5 Pro as its meta-agent but found it to perform worse, often yielding pipelines that fail due to creating plots or trying to evaluate code its evaluation agent would produce. For all methods the scaffold was limited to use up to 10 LLM calls per question.

Following ShinkaEvolve, each scaffold is evaluated on AIME 2024 three times. Each year the AIME competition consists of 30 questions, so this results in 90 question evaluations overall.

\section{Probabilities of Improvement and Dominance}
\label{app:prob_imp}
\textbf{Probability of Improvement.} Given methods $A,B$ that get scores $a_1,...,a_N$ and $b_1,...,b_N$ on some benchmark, following \citet{agarwal2021deep} the probability of improvement is
\begin{equation}
    P(A>B)=\frac{1}{N}\sum_{i=1}^NS(a_i,b_i) \quad\text{where}\quad S(a,b)=\begin{cases}
            1, & \text{if } a > b, \\
            \frac{1}{2}, & \text{if } a=b, \\
            0, & \text{if } a < b.
            \end{cases}
\end{equation}

Note that uncertainty over $P(A>B)$ can also be calculated using a bootstrap estimate. In cases where the probability of matching or exceeding is of interest, as in some parts of this work, the $\frac{1}{2}$ for when $a=b$ can be replaced with a $1$. See \citet{agarwal2021deep} for additional discussions.

\textbf{Probability of Dominance.} The probability of dominance is the probability method $A_1$ is better than (``dominates'') methods $A_2,A_3,...,A_M$. We denote their scores as $a_1^{(1)},a^{(1)}_2,...,a_N^{(1)},a_1^{(2)},a_2^{(2)},...$, with the upper index indicating the method. $P(A_1 > A_2, A_3, \dots, A_M)$ is defined as
\begin{equation}
P(A_1 > A_2, A_3, \dots, A_M)
=
\frac{1}{N^M}
\sum_{a^{(1)} \in A_1}
\cdots
\sum_{a^{(M)} \in A_M}
S\!\left(a^{(1)},\dots,a^{(M)}\right),
\end{equation}
where
\begin{equation}
S(a^{(1)},\dots,a^{(M)}) =
\begin{cases}
1, 
& \text{if } a^{(1)} > \max_{m \ge 2} a^{(m)}, \\[6pt]
\dfrac{1}{\left|\{m \ge 2 : a^{(m)} = a^{(1)}\}\right|},
& \text{if } a^{(1)} = \max_{m \ge 2} a^{(m)}, \\[10pt]
0,
& \text{otherwise.}
\end{cases}
\end{equation}
The second case in $S$ means that success probabilities are evenly split across the top methods in the case of ties. We refrain from the notation $P(A_1>\max(A_2,...,A_M))$ as the probabilities are calculated over the empirical distributions, not summary statistics. Although illustrated here for $A_1$ versus $A_2,...,A_M$ note that the method ordering is arbitrary. 

Although it is expensive to calculate the probability of dominance exactly, it can be efficiently estimated using Monte-Carlo. Comparing more methods will lead any individual method's probability of dominance to generally be lower, with several top methods likely having similar probabilities.

\textbf{Using Probability of Dominance with Evaluation Cascades.} We refrain from giving exact prescriptions as to the number of questions and cascades as what is sufficient is likely case dependent, noting that 300 questions is a sensible minimum for an expensive evaluation. As the effects of stochasticity worsen as more scaffolds are evaluated, stricter cascades should be used, using independent reruns to ensure an agent's performance is not spurious.

\section{MLE Bench Technical Details}
\label{app:mle_bench_details}
 We use AIDE prompts for the baselines, removing references to memory (previous programs) and using a single trial for SCS due to the time limit. All methods use Gemini-2.5 Pro and rely on the AIDE and MLE bench setups given in \citet{toledo2025ai}.

Notably, due to the competitions' intricacy, zero-shot generated programs almost never run without errors. This is due to there being many minute details in the dataset or the environment that might not be mentioned in the competition's description, e.g. very few samples in some classes leading models to fail when expecting at least a certain number of data points. Thus, we add a step to both baselines, where after generating a solution they iteratively debug it until it runs successfully. This results in slightly more complicated baselines and demonstrates a domain in which textual feedback is evidently important for functional code evolution, unlike the previous two domains. Unlike AIDE, the baselines do not include explicit fitness-based selection or a memory of all past programs and their performance.

\subsection{Full MLE Bench Competition Names}
\label{app:comp_names}
Table \ref{tab:comp_names} lists the unabbreviated names of the competitions in Table \ref{tab:mle_bench}.

\begin{table}[h]
    \centering
    \caption{Competition abbreviations and full names.}
    \begin{tabular}{cc}
    Abbreviation & Full Competition Name \\
    \midrule
    Insult Detection
    & Detecting Insults in Social Commentary \\
    
    Dec.\ 2021 Tabular
    & Tabular Playground Series -- December 2021 \\
    
    Trans.\ Conductors
    & Nomad2018: Predict Transparent Conductors \\
    
    English Text Norm.
    & Text Normalization Challenge -- English Language \\
    
    May 2022 Tabular
    & Tabular Playground Series -- May 2022 \\
    
    Random Pizza
    & Random Acts of Pizza \\
    
    Spooky Author
    & Spooky Author Identification \\
    
    Toxic Jigsaw
    & Jigsaw Toxic Comment Classification Challenge \\
    
    Russ.\ Text Norm.
    & Text Normalization Challenge -- Russian Language \\
    
    NYC Taxi
    & New York City Taxi Fare Prediction \\
\end{tabular}
    \label{tab:comp_names}
\end{table}

\subsection{Human Baseline Metrics}
\label{app:mle_human_metrics}
Table \ref{tab:mle_bench_human} shows the human baseline metrics used to calculate reference medal and median scores in Table \ref{tab:mle_bench}.

\begin{table*}
    \centering
    \caption{Human baseline metrics on the Kaggle competitions. Median is the median score in the competition, with bronze, silver, and gold denoting the cutoff score to receive each medal. Metrics are from \citet{schmidgall2025agent}.}
    \label{tab:mle_bench_human}
    \begin{tabular}{l|cccc}
        & \multicolumn{4}{c}{Human baseline metrics}
        \\
        Competition 
        & Median 
        & Bronze$^{\color{brown}*}$ 
        & Silver$^{\color{gray}+}$ 
        & Gold$^{\color{amber}\times}$ 
        \\
        \midrule
        Insult Detection \arrowtag{\uparrow}
        & 0.778 & 0.791 & 0.823 & 0.833
        \\

        Dec. 2021 Tabular \arrowtag{\uparrow}
        & 0.953 & 0.956 & 0.956 & 0.956
        \\

        Trans. Conductors \arrowtag{\downarrow}
        & 0.069 & 0.065 & 0.062 & 0.055
        \\

        English Text Norm. \arrowtag{\uparrow}
        & 0.990 & 0.990 & 0.991 & 0.997
        \\

        May 2022 Tabular \arrowtag{\uparrow}
        & 0.972 & 0.998 & 0.998 & 0.998
        \\

        Random Pizza \arrowtag{\uparrow}
        & 0.599 & 0.692 & 0.724 & 0.979
        \\

        Spooky Author \arrowtag{\downarrow}
        & 0.418 & 0.293 & 0.269 & 0.165
        \\

        Toxic Jigsaw \arrowtag{\uparrow}
        & 0.980 & 0.986 & 0.986 & 0.987
        \\

        Russ. Text Norm. \arrowtag{\uparrow}
        & 0.975 & 0.975 & 0.982 & 0.990
        \\

        NYC Taxi \arrowtag{\downarrow}
        & 3.597 & 2.923 & 2.881 & 2.337
        \\
    \end{tabular}
\end{table*}

\section{Tips for Economic Code Evolution Development}
\label{app:cheap_code_evo}
Many of the experiments in this paper are expensive. To enable easier development of code evolution methods we discuss a few tips to help keep costs low.

First, when developing, one could use a cheaper LLM. For example, for diagnostics we used Gemini-2.5 Flash Lite or Flash instead of Pro and sampled fewer programs. Another option is to use open-source LLMs which allow self-hosting. E.g. \citet{toledo2025ai} use a Deepseek model instead of API calls, albeit chiefly to have a higher throughput due to API rate limits. However, open-source models have the clear downside of being generally less capable than various closed source counterparts, and require renting GPUs if none are available. Still, capable open-source models are estimated to cost $5-15\times$ less than frontier closed-source models \citep{benchlm_llm_pricing_2026}.

Second, there are ways to make expensive evaluations cheaper. The evaluation cascade described in §\ref{subsec:stoch_red} helps keep the cost of agent evaluations low while still reducing stochasticity. \citet{schmidgall2025agent} run MLE bench experiments on laptop CPUs, thereby requiring only API calls and no GPUs. Although this limits their available toolset, it still allows comparing different pipelines under identical settings.

\section{Additional Limitations}
\label{app:limits}

Magnitudes of improvements are not compared for the mathematical bounds and machine learning competitions' results as small magnitudes can be meaningful, making it unclear how to effectively compare them. For example, the difference between the SCS and ShinkaEvolve circle packing bounds in Table \ref{tab:results_math} is on the order of $10^{-4}$. This difference is meaningful when it is close to the best bound found but generally meaningless otherwise. Instead, we opt to use pairwise comparisons as they can be meaningfully interpreted, with the downside of results over individual tasks being less meaningful.

Specifically for the MLE bench experiments, we note that although AIDE is more complicated than the baselines, it is still simpler than some other code evolution systems. We are unaware of a code evolution method that is both more complicated and more performant on MLE bench, and compare to AIDE due to it being both well known and performant.

\end{document}